\documentclass[final]{cvpr}

\usepackage{times}
\usepackage{graphicx}
\usepackage{amsmath}
\usepackage{amssymb}
\usepackage{overpic}
\usepackage{multirow}
\usepackage{makecell}
\usepackage{booktabs}
\usepackage{subfig}
\usepackage{cite}
\usepackage{pifont}
\newcommand{\cmark}{\ding{51}}%
\newcommand{\xmark}{\ding{55}}%
\DeclareGraphicsExtensions{.pdf,.jpg,.png}

\usepackage{enumitem}
\setenumerate[1]{itemsep=0pt,partopsep=0pt,parsep=\parskip,topsep=0pt}
\setitemize[1]{itemsep=0pt,partopsep=0pt,parsep=\parskip,topsep=0pt}
\newcommand{\tabincell}[2]{\begin{tabular}{@{}#1@{}}#2\end{tabular}}
\graphicspath{{./pics/}}

\usepackage[pagebackref=true,breaklinks=true,colorlinks,bookmarks=false]{hyperref}

\newcommand{\ConfInf}{\vspace{-.7in} {\normalsize \normalfont \color{blue}{
	IEEE International Conference on Computer Vision and Pattern Recognition (CVPR 2021)}} \vspace{.45in} \\}

\def\cvprPaperID{805} 
\def\confYear{CVPR 2021}

\begin{document}
	
	\title{\ConfInf DOTS: Decoupling Operation and Topology in Differentiable \\ Architecture Search}
	
	\author{Yu-Chao Gu$^{1*}$ \quad Li-Juan Wang$^1$\thanks{Both authors contributed equally to this work.} \quad 
		Yun Liu$^1$ \quad Yi Yang$^2$ \quad Yu-Huan Wu$^1$ \quad \\
		Shao-Ping Lu$^1$\quad Ming-Ming Cheng$^1$\thanks{M.M. Cheng (cmm@nankai.edu.cn) is the corresponding author.} \\
		$^1$TKLNDST, CS, Nankai University \hspace{.2in}
		$^2$Zhejiang University
	}
	
	\maketitle
	
	\begin{abstract}
		Differentiable Architecture Search (DARTS) has attracted extensive attention 
		due to its efficiency in searching for cell structures. 
		DARTS mainly focuses on the operation search 
		and derives the cell topology from the operation weights. 
		However, the operation weights can not indicate the importance of cell topology 
		and result in poor topology rating correctness.
		To tackle this, we propose to \textbf{D}ecouple the \textbf{O}peration and \textbf{T}opology \textbf{S}earch (DOTS),  
		which decouples the topology representation from operation weights 
		and makes an explicit topology search.
		DOTS is achieved by introducing a topology search space 
		that contains combinations of candidate edges. 
		The proposed search space directly reflects the search objective 
		and can be easily extended to support a flexible number of edges in the searched cell.
		Existing gradient-based NAS methods can be incorporated into DOTS 
		for further improvement by the topology search. 
		Considering that some operations (\textit{e.g.}, Skip-Connection) can affect the topology, 
		we propose a group operation search scheme 
		to preserve topology-related operations for a better topology search.
		The experiments on CIFAR10/100 and ImageNet demonstrate that DOTS is an effective solution for differentiable NAS.
		The code is released at \url{https://github.com/guyuchao/DOTS}.
	\end{abstract}
	\thispagestyle{empty}
	\section{Introduction}
	Neural Architecture Search (NAS) has attracted extensive attention  
	for its potential to find the optimal architecture 
	in a large search space automatically.
	Previous reinforcement learning and evolutionary learning based approaches 
	\cite{zoph2018learning,liu2018hierarchical,real2019regularized} 
	require a full training process to validate the architecture performance, 
	consuming hundreds of GPU-days to search. 
	To reduce the search cost, one-shot methods 
	\cite{enas,dong2019one,chu2019fairnas,guo2019single} 
	adopt the weight sharing strategy,
	which trains the \textit{supernet} once and derives child architecture 
	performance from the supernet directly.
	Recent methods 
	\cite{xu2020pcdarts,chen2019progressive,cai2018proxylessnas,xie2018snas} 
	based on differentiable architecture search (DARTS) \cite{liu2018darts} 
	also adopt the weight sharing strategy and further reduce the search cost 
	by unifying the supernet training and child architecture searching.

\begin{figure*}[t]
		\centering
		\subfloat[Operation Weight, CIFAR10]{\label{fig:rank_darts_c10}
			\includegraphics[width=.45\linewidth]{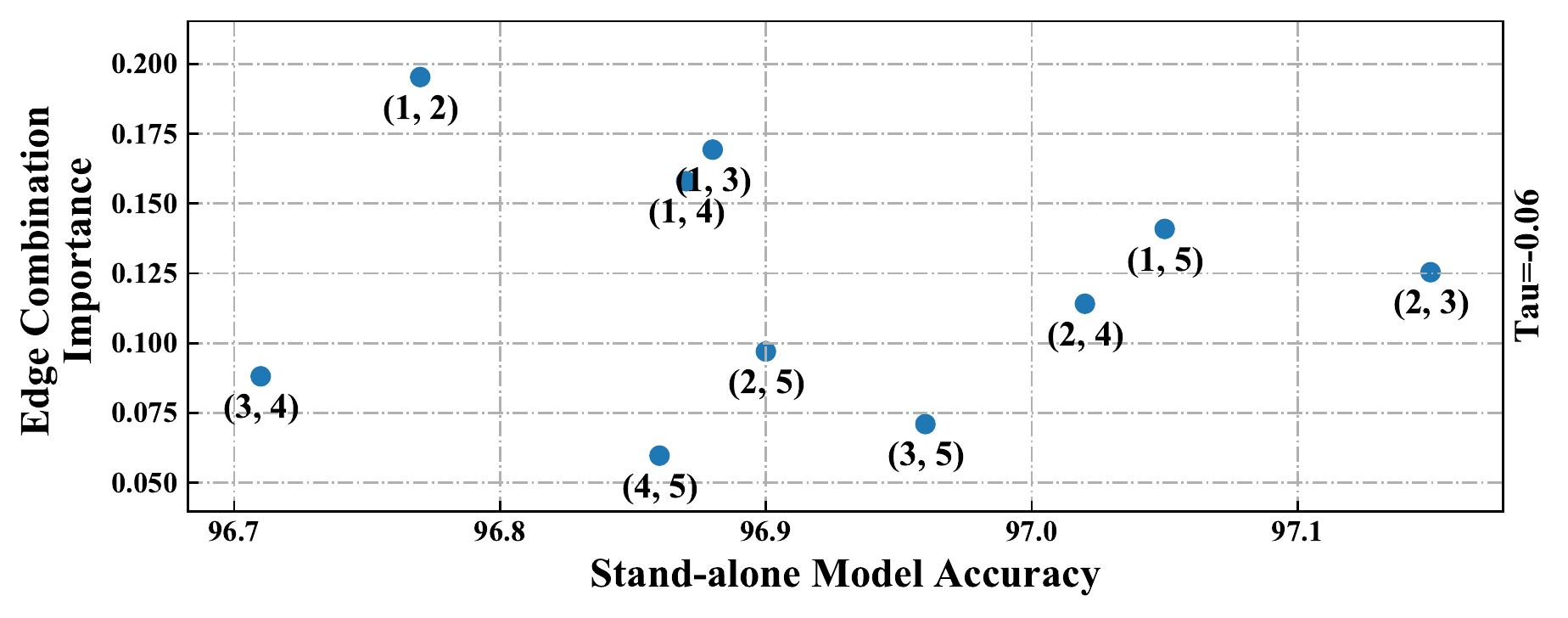}
		}
		\subfloat[Edge Combination Weight, CIFAR10]{\label{fig:rank_dots_c10}
			\includegraphics[width=.45\linewidth]{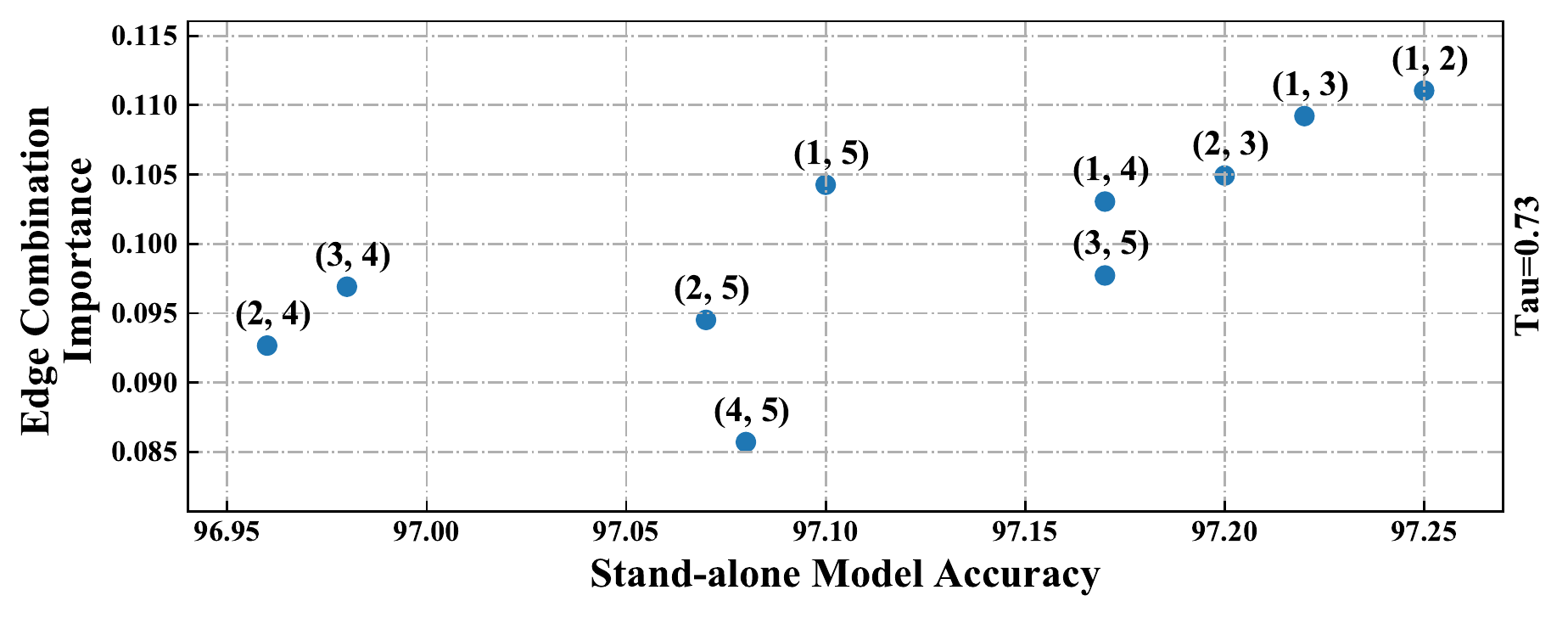}
		} \\
		\subfloat[Operation Weight, CIFAR100]{\label{fig:rank_darts_c100}
			\includegraphics[width=.45\linewidth]{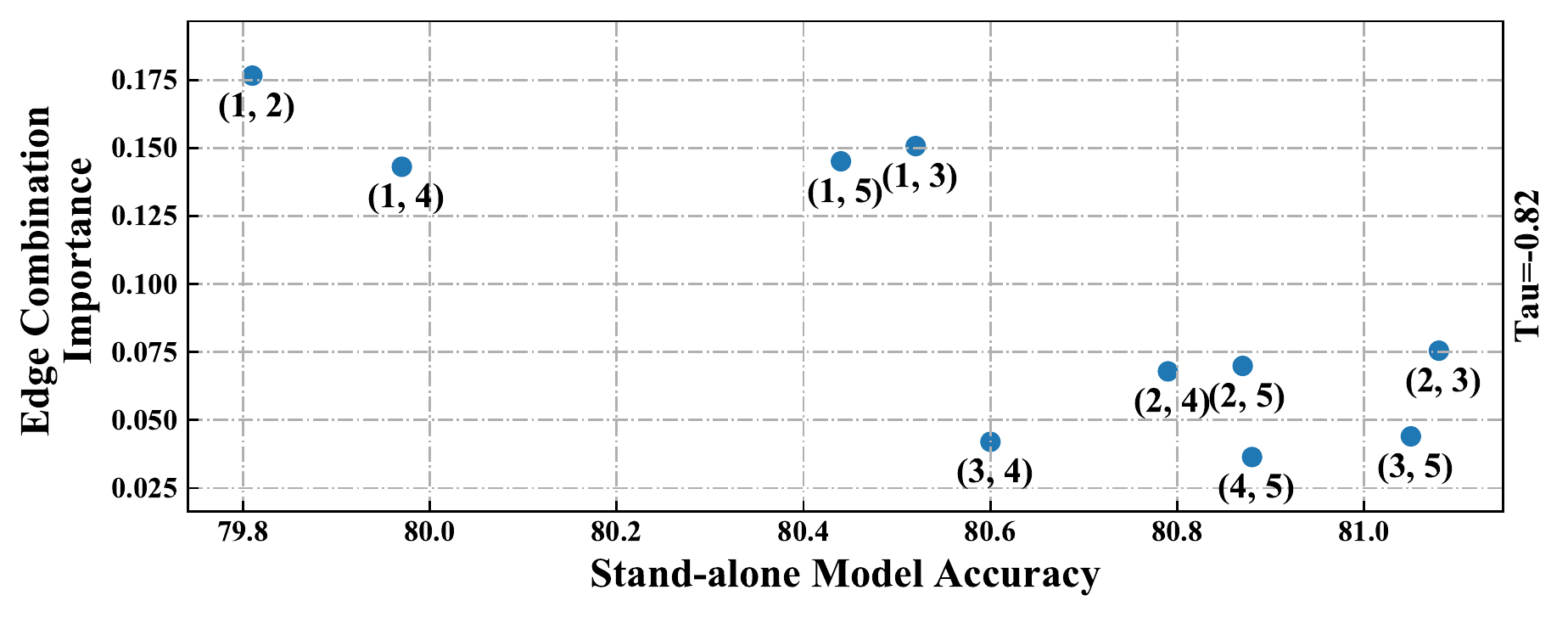}
		}
		\subfloat[Edge Combination Weight, CIFAR100]{\label{fig:rank_dots_c100}
			\includegraphics[width=.45\linewidth]{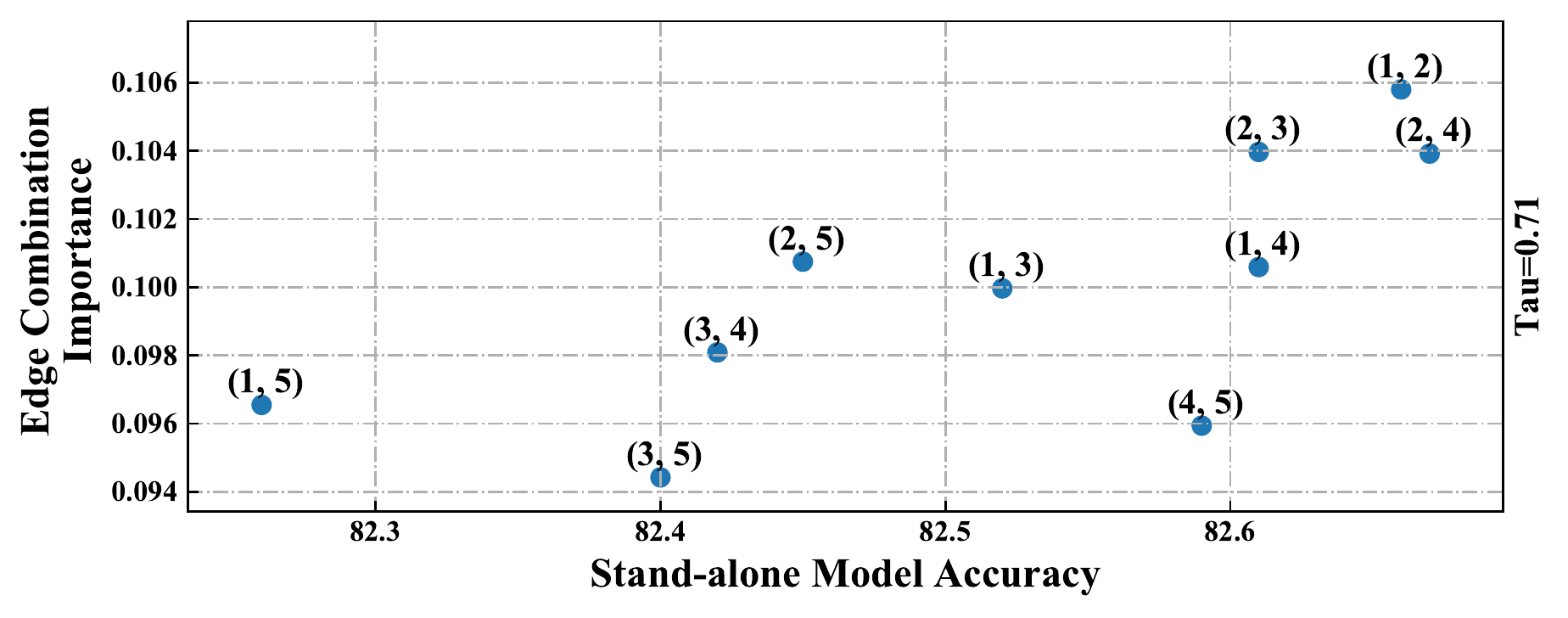}
		}
		\vspace{-.1in}
		\caption{Rank correlation analysis between different edge importance 
			representations and the stand-alone model performance.
			The edge combination importance is indicated by operation weights 
			(DARTS) and edge combination weights (DOTS). 
			We calculate the \textit{Kendall Tau} metric \cite{kendall1938new} 
			to measure the rank correlation.
		}\label{fig:otprank}
	\end{figure*}

	In DARTS, the operation selection is parameterized with learnable operation 
	weights, which is updated with the supernet training. 
	After training, the operation weights are used to rank the importance 
	of operations and topology. 
	The edge importance in DARTS is represented as the largest operation weight 
	on this edge.
	DARTS retains the two most important edges for each intermediate node 
	to derive the topology of the searched cell. 
	A question is raised: whether the edge importance indicated by the 
	operation weights accurately ranks the stand-alone model's performance.
	As illustrated in \figref{fig:otprank}, we find no obvious rank correlation, 
	which implies that DARTS has no superiority over choosing edges randomly 
	(see more details in \secref{sec:motivation}).  
	Furthermore, DARTS' handcraft policy of edge numbers restricts their potential 
	to find more flexible cell structures.	
	
	This paper addresses the above problems
	via \textbf{D}ecoupling \textbf{O}peration and \textbf{T}opology \textbf{S}earch (DOTS). 
	The meaning of decoupling is two-fold.
	On the one hand, we decouple the topology representation from the operation weights. 
	In detail, we introduce a topology search space containing the pairwise combinations of edges. The topology search space is continuously relaxed, and the relaxed topology weights model the combinatorial distribution of candidate edges. 
	The proposed topology search space directly reflects the search objective and can be easily extended to support a flexible number of edges.
	On the other hand, we decouple the operation and topology search processes. The overall search process is divided into the operation search stage and the topology search stage, 
	in which we update operation weights and edge combination weights, respectively. 
	With decoupling the two searching processes, existing gradient-based NAS methods can be directly incorporated into the DOTS' operation search and get further improvement by the topology search. Furthermore, the topology search is performed in a shrunk supernet, making it more efficient and accurate.
	Considering that some operations (\textit{e.g.}, \textit{Skip-Connection}) can affect the topology, we adopt a group strategy in the operation search to preserve these topology-related operations for a better topology search.
	
	We summarize our contributions as follows:
	\begin{itemize}
		\item We propose to decouple the operation and topology search, which decouples both the topology representation and search processes. Such decoupling leads to the correct rating of stand-alone models with different topologies.  
		\item The proposed topology search space can be extended to support a flexible number of edges in the searched cell, fulfilling its potential to search for more complex structures.
		\item 
		Existing gradient-based methods can be incorporated into DOTS and get further improvement by the topology search. 
		DOTS only costs 0.26 and 1.3 GPU-days to search from scratch and achieves 97.51\% and 76.0\% accuracy on CIFAR10 and ImageNet. Better performance can be achieved if the constraint of edge numbers is removed.
	\end{itemize}

	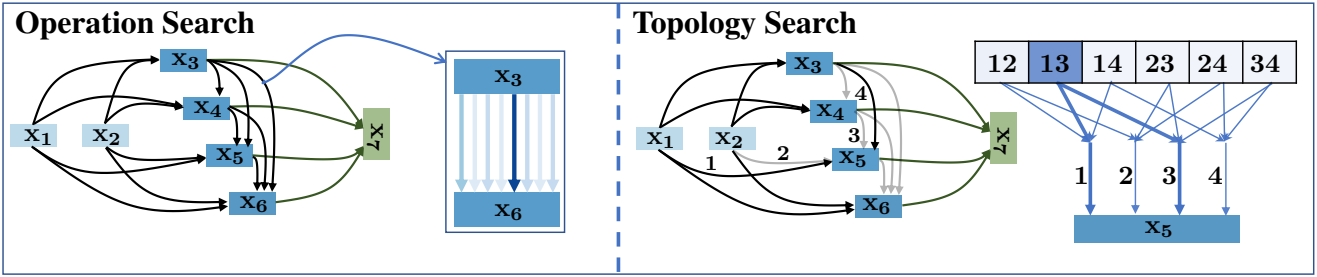
\begin{figure*}[t]
		\centering
		\begin{overpic}[width=\linewidth]{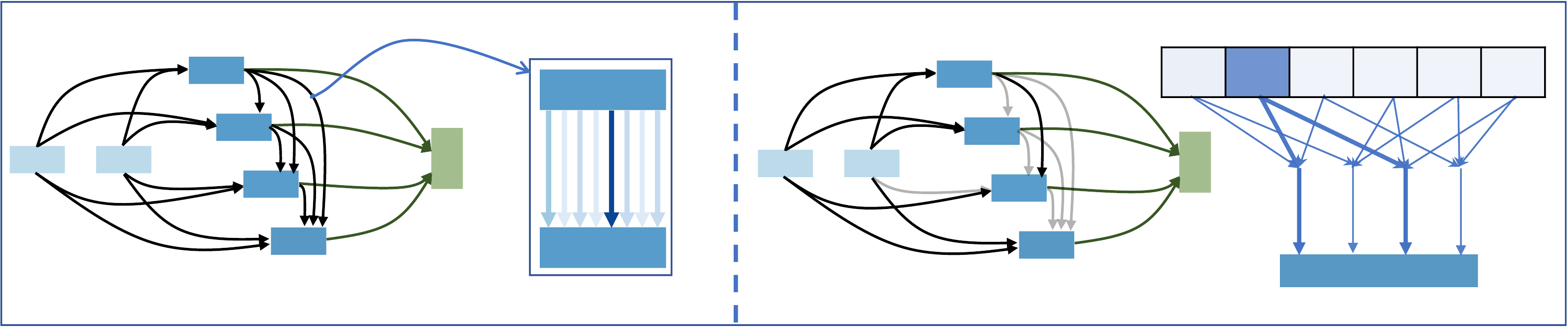}
			\put(1.7, 10.25){$\mathbf{x_1}$}
			\put(6.9, 10.25){$\mathbf{x_2}$}		
			\put(12.9, 16){$\mathbf{x_3}$}
			\put(14.6, 12.4){$\mathbf{x_4}$}
			\put(16.3, 8.8){$\mathbf{x_5}$}
			\put(18.1, 5.1){$\mathbf{x_6}$}
			\put(37.5, 14.7){$\mathbf{x_3}$}
			\put(37.5, 4.5){$\mathbf{x_6}$}		
			\put(27.8, 11.6){\rotatebox{-90}{$\mathbf{x_7}$}}
			
			\put(49.1, 10){$\mathbf{x_1}$}
			\put(54.3, 10){$\mathbf{x_2}$}		
			\put(60.3, 15.75){$\mathbf{x_3}$}
			\put(62, 12.15){$\mathbf{x_4}$}
			\put(63.7, 8.55){$\mathbf{x_5}$}
			\put(65.5, 4.95){$\mathbf{x_6}$}
			\put(75.5, 11.3){\rotatebox{-90}{$\mathbf{x_7}$}}
			\put(87,3.2){$\mathbf{x_5}$}
			\footnotesize
			\put(53.5, 8){$\mathbf{1}$}
			\put(59, 8.8){$\mathbf{2}$}		
			\put(64.4, 10.2){$\mathbf{3}$}
			\put(64.9, 13.4){$\mathbf{4}$}
			\normalsize
			\put(81.6, 7){$\mathbf{1}$}
			\put(85, 7){$\mathbf{2}$}		
			\put(88.3, 7){$\mathbf{3}$}
			\put(91.8, 7){$\mathbf{4}$}
			
			\put(75, 15.5){$\mathbf{12}$}
			\put(79, 15.5){$\mathbf{13}$}
			\put(83, 15.5){$\mathbf{14}$}
			\put(87, 15.5){$\mathbf{23}$}
			\put(91, 15.5){$\mathbf{24}$}
			\put(95, 15.5){$\mathbf{34}$}
			\large
			\put(1, 18.5){\textbf{Operation Search}}
			\put(48, 18.5){\textbf{Topology Search}}
		\end{overpic}
		\vspace{-.3in}
		\caption{Overall pipeline of the proposed DOTS. 
			The DOTS framework consists of the operation search and the topology search. In the operation search phase, we search for the best performing operations on each edge. And in the topology search phase, we search for the best combination for candidate edges.}
		\label{fig:pipeline}
	\end{figure*}
	
	\section{Related Work}
	
	Different from the previous manually designing task-specific neural networks \cite{gu2020pyramid,wu2021jcs,liu2020leveraging,wu2020regularized,gao2021rbn},  
	Neural Architecture Search (NAS) has attracted extensive attention   
	for its potential to design efficient networks automatically \cite{liu2019auto,ghiasi2019fpn,li2020neural,gao2021global2local}. 
	Early methods based on reinforcement learning \cite{zoph2016neural,zoph2018learning} 
	and evolutionary algorithms \cite{xie2017genetic,real2019regularized} train thousands of candidate architectures from scratch and use their validation accuracy to learn a meta-controller, which requires prohibitive search cost.
	Recent one-shot NAS methods \cite{brock2018smash,Bender18oneshot,guo2019single,chu2019fairnas} and gradient-based methods \cite{liu2018darts,chu2019fair,he2020milenas,xie2018snas} adopt the weight sharing strategy \cite{enas}, which only trains the \textit{supernet} once and thus reduces the search cost. 
	Recent gradient-based methods try to overcome the instability \cite{liang2019darts+,Zela2020Understanding,chu2019fair,he2020milenas,xie2018snas,chen2020stabilizing,chu2020darts} and reduce the search cost \cite{chen2019progressive,dong2019searching,xu2020pcdarts}. Previous gradient-based methods mainly target improving the operation search.
	While our work improves gradient-based methods with the topology search, which is complementary to previous researches.

	Recent researches \cite{xie2019exploring,Shu2020Understanding,fang2020densely,gao2019res2net} 
	reveal the importance of connection topology in neural networks. The randomly-wired network \cite{xie2019exploring} finds that networks generated by random graph algorithms  can obtain competitive results. Shu \textit{et al.} \cite{Shu2020Understanding} find that the cell topology has more impact on network convergence than the operation in cell-based NAS. DenseNAS \cite{fang2020densely} proposes a densely-connected search space that focuses on the macro structure's topology. Our work sheds light on the micro cell's topology and constructs a topology search space to support a flexible number of edges.
	
	Recent weight sharing methods \cite{li2020sgas,Yu2020Evaluating,li2020improving,li2020block,zheng2020rethinking,zhou2020econas,hu2020dsnas} try to improve the architecture rank correctness.
	Yu \textit{et al.} \cite{Yu2020Evaluating} point out that recent NAS has similar performance to random search because of the constrained search space and the widely-used weight sharing strategy.
	PCNAS \cite{li2020improving} identifies and fixes the posterior fading problem in weight-sharing methods. Block-wisely NAS \cite{li2020block} improves the architecture rank correctness by modularizing the large space into blocks. 
	Although some recent works notice the rating problem in one-shot methods, there is less concern about the rating problem in gradient-based methods.  

	\section{Review of DARTS}
	
	\subsection{Preliminary of DARTS}
	\label{sec:darts}
	We start by reviewing the baseline algorithm DARTS \cite{liu2018darts}. 
	DARTS aims at searching for the cell, 
	a repeating building block of the neural network. 
	A cell is represented as a directed cyclic graph with $N$ nodes $\{x_i\}^{N}_{i=1}$, 
	including two input, one output, and several intermediate nodes.
	Each node denotes a feature map transformed by graph edges.
	The $j$-th intermediate node $x_j$ connects to all its predecessors $x_i$ through the edge $(i,j)$.
	Each edge $(i,j)$ contains candidate operations weighted by the operation weight $\alpha^{(i,j)}$,
	which can be defined as
	
	\begin{equation}
	\label{eq:opbar}
	\bar{o}^{(i,j)}(x)=\sum_{o\in O}\alpha_o^{(i,j)}o^{(i,j)}(x_i),
	\end{equation}
	where $o(x)\in \mathcal{O}$ and $\mathcal{O}$ is the operation search space containing eight operations, including \textit{Zero}, \textit{Skip-Connection}, \textit{Avg-Pooling}, 
	\textit{Max-Pooling}, \textit{Sep3x3Conv}, \textit{Sep5x5Conv}, \textit{Dil3x3Conv}, and \textit{Dil5x5Conv}. 
	The weight for each operation is normalized with softmax:
	
	\begin{equation}
	\label{eq:opweight}
	\alpha_o^{(i,j)}=\frac{\exp({\alpha^{'}}_{o}^{(i,j)})}{\sum_{o^{'}\in \mathcal{O}}\exp({\alpha^{'}}_{o^{'}}^{(i,j)})},
	\end{equation}
	where $\alpha^{'}$ is the unnormalized operation weight.
	The operation weight is updated with the supernet training, 
	gradually focusing on the optimal architecture. 
	Once defined the mixed operation $\bar{o}^{(i,j)}$ for edge $(i,j)$,
	the intermediate node $x_j$ is computed from all its predecessors $x_i$:
	
	\begin{equation}
	x_j=\sum_{i<j}\bar{o}^{(i,j)}(x_i).
	\end{equation}
	Let $\mathcal{L}_{cls}^{train}$ and $\mathcal{L}_{cls}^{val}$ be the Cross-Entropy loss on the training and validation sets, respectively.
	Then, we can formulate the bi-level optimization with the operation weight $\alpha$ and the network weight $w$ as 
	
	\begin{equation}
	\begin{split}
	&\min_{ \alpha}\ \ \mathcal{L}_{cls}^{val}(w^{*}(\alpha),\alpha),\\
	{\rm s.t.}\ \ w^{*}(&\alpha)=\arg \min_{w}(\mathcal{L}_{cls}^{train}(w,\alpha)).
	\end{split}
	\end{equation}
	After searching, the final architecture is derived from the operation weight $\alpha$ by two hard pruning: 
	\begin{enumerate}
		\item Retain the operation with the largest weight and prune other operations for each edge, \textit{i.e.}, $o^{(i,j)}=\arg\max_{o\in\mathcal{O},o\neq Zero}\alpha_{o}^{(i,j)}$.
		\item Retain two incoming edges with the largest edge importance for each intermediate node and prune other edges.
		The edge importance is defined as the largest operation weight on each edge $(i,j)$, \textit{i.e.}, $\max_{o\in\mathcal{O},o\neq Zero}\alpha_{o}^{(i,j)}$.
	\end{enumerate}

	\subsection{Coupling Problem in DARTS}
	\label{sec:motivation}
	
	Previous works \cite{xie2018snas,chen2019progressive,xu2020pcdarts,dong2019searching,liang2019darts+,li2019stacnas,noy2020asap,chu2020darts,chen2020stabilizing,hong20drop} based on DARTS indicate the edge importance by the largest operation weight (excluding the \textit{Zero} operation) on this edge. 
	We conduct a rank correlation analysis to find whether edge importance indicated by operation weights can accurately rank the stand-alone model (measured by \textit{Kendall Tau} metric \cite{kendall1938new}). 
	We follow DARTS' handcraft policy that sums the importance of two edges to get its edge combination importance. 
	However, for DOTS, the edge combination weights can directly represent the edge combination importance.
	There are five edges in our experiment, resulting in ten different edge combinations. The stand-alone model is trained with the same setting as in \secref{sec:C10bench}, except that we reduce the number of training epochs to 300.  
	
	\figref{fig:rank_darts_c10} and \figref{fig:rank_darts_c100} show that the stand-alone model accuracy has no clear ranking correlation with edge importance indicated by the operation weights. It implies that DARTS' searched cell is sub-optimal because it cannot converge to the best cell topology, which is consistent with the finding that DARTS cannot surpass random search in some cases.
	Intuitively, the larger operation weight can only indicate an operation's suitability for a specific edge,
	but does not mean the edge should be retained in the searched topology. 
	As shown in \figref{fig:rank_dots_c10} and \figref{fig:rank_dots_c100}, the proposed edge combination weights achieve $Tau=0.73$ and $Tau=0.71$ on CIFAR10 and CIFAR100, demonstrating its effectiveness for edge selection.

	\section{Methodology}
	
	The above analysis points out the limitation of coupling the operation and topology search. In this section, we try to tackle this problem via decoupling operation and topology search. As shown in \figref{fig:pipeline}, the overall search is divided into the operation search and the topology search. In the operation search phase,  we search for the best operation on each edge. In the topology search phase, we search for the best combination of candidate edges. In \secref{sec:tpsearch}, we introduce how to construct the topology search space and support a flexible number of edges. In \secref{sec:opsearch}, we describe how to incorporate existing gradient-based NAS methods into DOTS' operation search and propose a group operation search scheme to retain topology-related operations for a better topology search.
	
	\subsection{Topology Search}
	\label{sec:tpsearch}
	
	\subsubsection{Handcraft Policy for the Number of Edges}
	\label{sec:handcraft}
	In \secref{sec:motivation}, we have discussed the limitation of coupling the operation and topology search.
	Hence, we need to decouple the edge importance from operation weights. 
	To achieve this, we define a topology search space apart from the operation search space. 
	
	Formally, the \textit{j}-th intermediate node $x_j$ connects to all its predecessors $x_i$ through the edge $(i,j)$. Following DARTS' handcraft policy to restrict two edges for intermediate nodes, we make the topology search space $\mathcal{E}_{x_j}$ for $x_j$ as all pairwise combinations of its incoming edges:
	
	\begin{equation}
	\label{eq:tpsearchspace}
	\mathcal{E}_{x_j}=\{\left<(i_1,j),(i_2,j)\right>|0<i_1<i_2<j \}.
	\end{equation}
	Suppose that there are total $n$ incoming edges for node $x_j$, so the search space  $\mathcal{E}_{x_j}$ contains $C_{n}^{2}=\frac{n!}{2!(n-2)!}$ candidate combinations. 
	For each node $x_j$, we relax its topology search space to continuous, which can be defined as
	
	\begin{equation}
	\label{eq:tprelaxation}
	\beta^{c}_{x_j}=\frac{\exp({\beta^{'}}_{c}^{x_j}/T_{\beta})}{\sum_{c^{'}\in \mathcal{E}_{x_j}}\exp({\beta^{'}}_{c^{'}}^{x_j}/T_{\beta})},
	\end{equation}
	where $\beta^{c}_{x_j}$ denotes the normalized probability of choosing edge combination $c$. Although the topology search space is defined on edge combinations, we do not need to obtain each edge combination feature in practice. To reduce the memory cost, we aggregate weight for edge $e_j^i$ from those combinations containing this edge, which can be formulated as
	
	\begin{equation}
	\label{eq:aggregate}
	\gamma^{(i,j)}=\sum_{c\in \mathcal{E}_{x_j},(i,j)\in c} \frac{1}{N(c)}\beta^{c}_{x_j},
	\end{equation}
	where $\gamma^{(i,j)}$ is the weight of each edges and $N(c)$ is the edges number in edge combination $c$. 
	We sum all the incoming edges of $x_j$ weighted by the edge importance weight $\gamma$ to get its features:
	
	\begin{equation}
	\label{eq:tpmixop}
	x_j=\sum_{i<j}\gamma^{(i,j)} \cdot \bar{o}^{(i,j)}(x_i),
	\end{equation}
	where $\bar{o}^{(i,j)}$ means the mixed operation on edge $(i,j)$. 
	Since we decouple the operation and topology search processes, the $\bar{o}^{(i,j)}$ mixes the candidate operations retained by the operation search, which will be discussed in \secref{sec:opsearch}.
	
	As discussed in ASAP \cite{noy2020asap} and SNAS \cite{xie2018snas}, the optimization gap between the supernet and the derived child network causes a performance drop. Both works exploit architecture annealing to bridge the optimization gap during searching. We generalize the annealing idea to the topology search. In \equref{eq:tprelaxation}, $T_{\beta}$ is the annealing temperature. We adopt an exponential schedule for annealing:
	
	\begin{equation}
	\label{eq:tpanneal}
	T(t)=T_0\theta^{t},
	\end{equation}
	where it starts from an initial temperature $T_0$ and decays with the training step $t$ increasing.
	
	DARTS uses bi-level optimization to avoid overfitting \cite{liu2018darts}.
	However, \cite{li2019stacnas,guo2019single} shows that one-level optimization is stable and accurate.
	In our topology search stage, the operation on each edge is largely reduced, eliminating the risk of overfitting.
	Therefore, we use one-level optimization for updating the network weight $w$ and the topology weight $\beta$, 
	which can be formulated as
	
	\begin{equation}
	\begin{split}
	w_t&=w_{t-1}-\eta_{t}\partial_{w}L_{train}(w_{t-1},\beta_{t-1}),\\
	\beta_{t}&=\beta_{t-1}-\delta_{t}\partial_{w}L_{train}(w_{t-1},\beta_{t-1}),
	\end{split}
	\end{equation}
	where $\eta_{t}$ and $\delta_{t}$ are the learning rates of the network weight and topology weight, respectively.
	
	\subsubsection{Flexible Number of Edges}
	\label{sec:flexible_numbers}
	In \secref{sec:handcraft}, we construct a topology search space following the DARTS handcraft policy to restrict each intermediate node in the searched cell connects two edges. 
	Such a handcraft policy cannot learn the number of edges automatically. 
	Hence, we extend the topology search space to support arbitrary numbers of edges. 
	Specifically, we adopt binary code to describe such search space.
	For the intermediate node $x_j$ with $n$ incoming edges, the binary code for the $m$-th edge combination can be represented as  
	
	\begin{equation}
	c_{m}=\{e_1,e_2,\dots,e_n\},
	\end{equation}
	where $e_i\in \{0,1\}$ denotes the edge $i$ exists or not in an edge combination. The topology search space for the node $x_j$ can be defined as
	\begin{equation}
	\label{eq:tpnewspace}
	\mathcal{E}_{x_j}=\{c^1,c^2,\dots,c^M \},
	\end{equation}
	where $M$ is the number of valid edge combinations. 
	We can compute $M$ by
	\begin{equation}
	M=\sum_{k=1}^{n} C_{n}^{k}=2^n-1.
	\end{equation}
	We exclude the extreme case where all edges are not in the edge combination.
	Overall, we enable searching for a flexible number of edges.
	This is easy to implement as we only replace the search space of \equref{eq:tpsearchspace} with \equref{eq:tpnewspace} and keep the architecture relaxation and optimization the same.

	\subsection{Operation Search}
	\label{sec:opsearch}
	
	In this section, we introduce the operation search of DOTS.
	The operation search aims to retain the optimal operation candidate on each edge.
	We introduce how to incorporate the existing gradient-based NAS methods into DOTS' operation search in \secref{sec:cellbased}. 
	Existing methods only retain the best operation on each edge, discarding that some operations (\textit{e.g.}, \textit{Skip-Connection}) may affect the topology. 
	To prevent topology-related operations from being pruned in the operation search, we propose a group operation search scheme in \secref{sec:groupbased}.

	\begin{figure}[!tb]
		\centering
		\includegraphics[width=.9\linewidth]{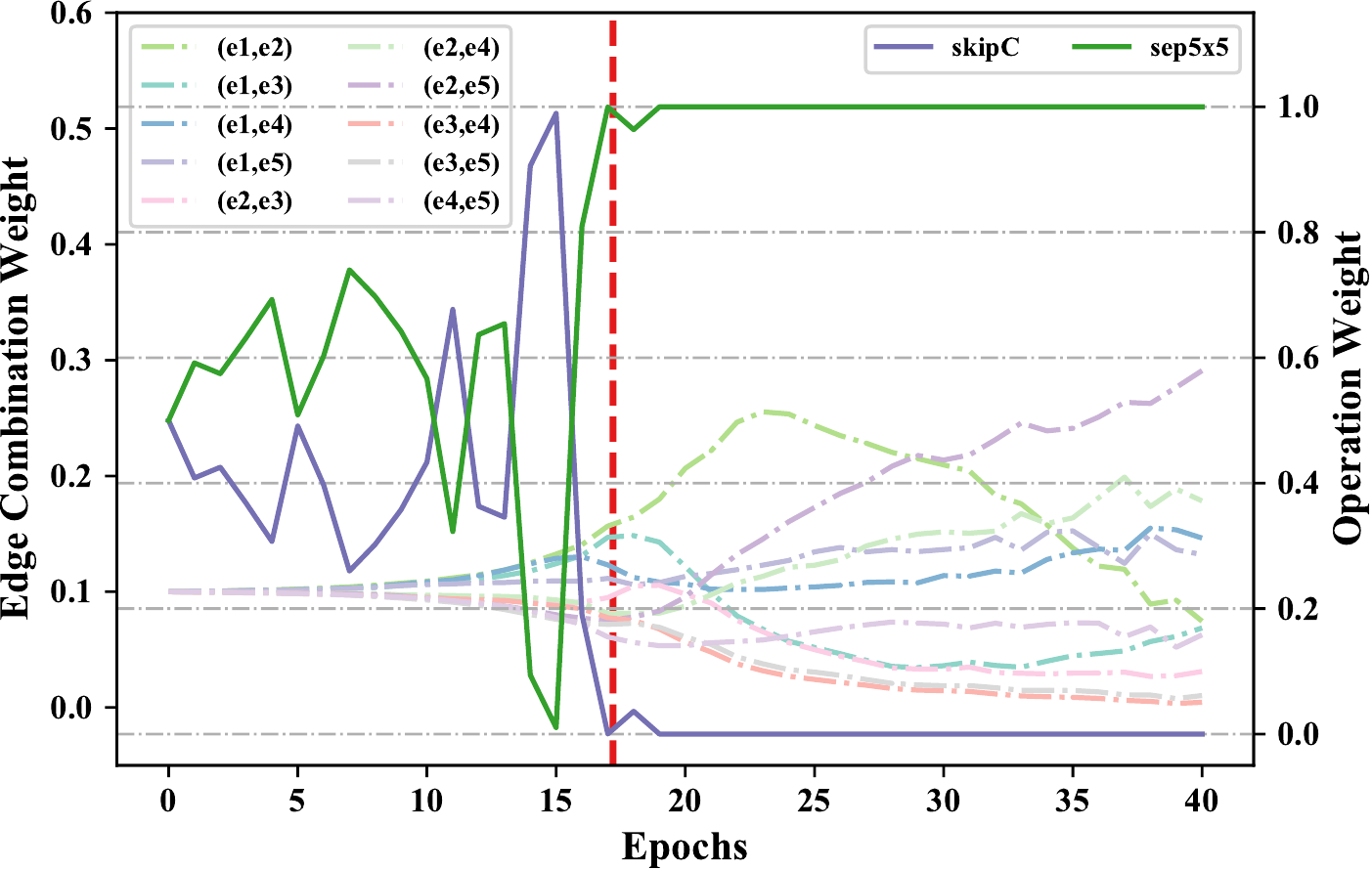}
		\vspace{-.1in}
		\caption{Topology search with more than one operation on each edge. We adopt a lower initial temperature to anneal operation weight than topology weight, \textit{i.e.}, $T_{\alpha_{o_n}}=\frac{1}{1000}T_{\beta}$. Hence, the operation is fixed on each edge in the first few epochs (denoted by the red dash line).  }
		\label{fig:anneal}
	\end{figure}
	
	\subsubsection{Incorporating Gradient-Based NAS Methods}
	\label{sec:cellbased}
	The previous gradient-based methods based on DARTS can be easily incorporated into DOTS' operation search. 
	DARTS introduces a set of weights $\alpha=\{\alpha^{(i,j)}\}$ for candidate operations on each edge $(i,j)$. We follow their own searching strategies to get the trained operation weight $\alpha$. Then, the operation with the maximum weight is retained on each edge $(i,j)$:
	\begin{equation}
	o^{(i,j)}=\arg\max_{o\in\mathcal{O}}\alpha_{o}^{(i,j)}.
	\end{equation}
	The final step is to replace the mixed operation $\bar{o}^{(i,j)}$ in \equref{eq:tpmixop} with $o^{(i,j)}$. 
	Overall, existing gradient-based NAS methods can be easily plugged into DOTS' operation search and get further improvement by the topology search.
	
	\subsubsection{Operation Search with Group Strategy}
	\label{sec:groupbased}
	
	Generally, the operations can be categorized into topology-related (\textit{e.g.}, \textit{Skip-Connection}) and topology-agnostic (\textit{e.g.}, \textit{Separable Convolution}). 
	In \secref{sec:cellbased}, we incorporate the existing gradient-based methods into DOTS' operation search, where the best operation is retained on each edge. 
	Such a strategy eliminates potential topology choices because some topology-related operations are dropped before the topology search. To this end, we resort to group strategy \cite{li2019stacnas,hong20drop} for the operation search.
	The group strategy can ensure to retain both topology-related and topology-agnostic operations for the topology search.
	
	Formally, in the group operation search, 
	the operation search space $\mathcal{O}$ is divided into several subspaces $\mathcal{O}=\{\mathcal{O}_1, \mathcal{O}_2, \dots \mathcal{O}_p\}$, where $p$ is the number of groups. Each operation subspace is relaxed to continuous independently.
	After searching, the operation with the largest weight is chosen from each group to construct a new operation search space $\mathcal{O}_{n}$ on each edge $(i,j)$, which can be formulated as
	
	\begin{equation}
	o_p^{(i,j)}=\arg\max_{o_p\in\mathcal{O}_p}\alpha_{o_p}^{(i,j)},
	\end{equation}%
	\begin{equation}
	\label{eq:newop}
	\mathcal{O}_{n}=\{o^{(i,j)}_1,o^{(i,j)}_2,\dots,o^{(i,j)}_p\}.
	\end{equation}%

	We evaluate different group strategies and discuss them in the experiment. 
	Since $O_n$ contains more than one operation, we need to search for it in the topology search stage. Simultaneously updating the operation and topology is inaccurate because it increases the weight-sharing child models \cite{chu2019fairnas,li2020improving}. To tackle this problem, we anneal the operation weight with a lower temperature $T_{\alpha_{o_n}}$ than $T_{\beta}$ in the topology search. 
	As illustrated in \figref{fig:anneal}, the operations in two groups are fixed in the first few epochs. Once the operations are fixed, further topology search evolves similarly as that in \secref{sec:cellbased}.

	\begin{table*}[!tb]
		\centering
		\small
		\setlength{\tabcolsep}{3.0mm}
		\resizebox{0.8\linewidth}{!}{%
			\begin{tabular}{lcccccc}
				\toprule[1pt]
				\textbf{Architecture} & \tabincell{c}{\textbf{Top-1 Acc. (\%)}\\ \textbf{CIFAR10}} & \tabincell{c}{\textbf{Params (M)}\\\textbf{CIFAR10}} & \tabincell{c}{\textbf{Top-1 Acc. (\%)}\\ \textbf{CIFAR100}} & \tabincell{c}{\textbf{Params (M)}\\\textbf{CIFAR100}} & \tabincell{c}{\textbf{Search Cost}\\\textbf{(GPU-days)}} & \textbf{Search Method} \\ \midrule[0.5pt]
				DenseNet-BC \cite{huang2017densely} &  96.54 & 25.6 &  82.82$^{\dag}$  & 25.6   & N/A & N/A  \\ \midrule[0.5pt]
				NASNet-A  \cite{zoph2018learning} &  97.35  & 3.3 &  83.18 & 3.3   & 1800 & RL  \\
				ENAS \cite{enas} &  97.11 & 4.6 & 80.57$^{\dag}$  & 4.6   & 0.5 & RL  \\ 
				AmoebaNet-B  \cite{real2019regularized} &  97.45$\pm$ 0.05   & 2.8  & - & -  & 3150 & EA  \\
				Hireachical Evolution \cite{liu2018hierarchical} &  96.25$\pm$ 0.12   & 15.7 & - & -  & 300 & EA  \\
				PNAS \cite{liu2018progressive} &  96.59$\pm$ 0.09  & 3.2 & 80.47$^{\dag}$  & 3.2  & 225 & SMBO  \\ \midrule[0.5pt]
				DARTS \cite{liu2018darts} &  97.00 & 3.4  & 82.46$^{\dag}$   &  3.4  & 0.4   & GD  \\
				SNAS \cite{xie2018snas} &  97.15   & 2.8 & 82.45 & 2.8  & 1.5  & GD  \\
				GDAS \cite{dong2019searching} &  97.07 & 2.5 & 81.62$^{\dag}$   & 3.4  & 0.2   & GD  \\
				P-DARTS \cite{chen2019progressive} &  97.50  & 3.4  & 82.51$^{\dag}$ & 3.6  & 0.3  & GD  \\ 
				FairDARTS \cite{chu2019fair} &  97.46 & 2.8 & 82.39 & 2.8 & 0.4  & GD \\
				PC-DARTS \cite{xu2020pcdarts} &  97.43 $\pm$ 0.07  & 3.6 & 83.10 & 3.6  & 0.1  & GD  \\
				DropNAS \cite{hong20drop} &  97.42 $\pm$ 0.14  & 4.1  & 83.13 & 4.0  & 0.6  & GD \\
				MergeNAS \cite{wang2020mergenas} &  97.27 $\pm$ 0.02  & 2.9 & 82.42 & 2.9  & 0.2  & GD\\
				ASAP \cite{noy2020asap} &  97.32 $\pm$ 0.11  & 2.5 & 82.69 & 2.5  & 0.2  & GD \\
				SDARTS-ADV \cite{chen2020stabilizing} &  97.39 $\pm$ 0.02  & 3.3 & 83.27 & 3.3  & 1.3  & GD\\
				DARTS- \cite{chu2020darts} &  97.41 $\pm$ 0.08 & 3.5 & 82.84$^{\dag}$ & 3.4  & 0.4  & GD\\ \midrule[0.5pt]
				DOTS (best) & 97.63 & 3.5 &  83.72  & 4.1 & 0.26 & GD  \\
				DOTS (avg)$^*$ & 97.51$\pm$0.06 & 3.5 &  83.52$\pm$0.13  & 4.1 & 0.26 & GD  \\
				\bottomrule[1pt]
		\end{tabular}}
		\vspace{-.13in}
		\caption{Comparison with state-of-the-art models on CIFAR10/100. CIFAR10 evaluation results are taken from their original paper.  $^{\dag}$: The CIFAR100 results are reported by Chu et al \cite{chu2020darts}. $^*$: Our results are obtained by four individual runs of search and evaluation under different random seeds.}
		\label{tab:CIFAR10}
	\end{table*}
	
	\section{Experiment}
	
	\subsection{Evaluation on CIFAR10/100}
	\label{sec:C10bench}
	
	\begin{table}[!tb] 
		\centering
		\small
		\resizebox{0.75\linewidth}{!}{%
		\begin{tabular}{l | c | c c c}
			\toprule[1pt]
			\textbf{Architecture}	&    \textbf{TS}     & \textbf{CIFAR10} & \textbf{CIFAR100}   \\ \midrule[0.5pt]
			\multirow{2}{*}{DARTS \cite{liu2018darts}} & \xmark & 97.02$\pm$0.12 &  80.74\\
			& \cmark & 97.40$\pm$0.09 & 83.07 \\ \midrule
			\multirow{2}{*}{DARTS (2nd) \cite{liu2018darts}} & \xmark & 97.01$\pm$0.15 &81.37  \\
			& \cmark &97.12$\pm$0.11 & 83.28 \\ \midrule
			\multirow{2}{*}{GDAS \cite{dong2019searching}} & \xmark &96.84$\pm$0.06 & 82.75\\
			& \cmark &97.06$\pm$0.08 & 83.01 \\ \midrule
			\multirow{2}{*}{SNAS \cite{xie2018snas}} & \xmark &97.05$\pm$0.10 & 81.92\\
			& \cmark &97.26$\pm$0.12 & 83.25 \\ \midrule
			\multirow{2}{*}{PC-DARTS \cite{xu2020pcdarts}} & \xmark & 97.28$\pm$0.08& 81.74\\
			& \cmark & 97.45$\pm$0.06&82.36  \\
			\bottomrule[1pt]
		\end{tabular}}
		\vspace{-.12in}
		\caption{Improving existing gradient-based NAS methods by the topology search. TS means the topology search.}
		\vspace{-.1in}
		\label{tab:improve_existing}
	\end{table}

	\textbf{Search Settings.}
	We implement the DOTS based on Pytorch \cite{paszke2019pytorch}, Mindspore \cite{mindspore} and Jittor \cite{hu2020jittor} frameworks.
	The whole search process on CIFAR10/100 takes 70 epochs, \textit{i.e.},
	30 epochs for the operation search and 40 for the topology search.
	The network is composed of 8 cells for the operation search and 20 cells for the topology search. 
	The initial temperature is set to $T_0=10$ and decay to 0.02 in the topology search. 
	The search process costs 6.3 hours (0.26 GPU-days) on one NVIDIA Tesla V100 GPU.
	More detailed search settings can be found in the supplementary.

	\textbf{Evaluation Settings.}
	The evaluation network is composed of 20 cells (18 normal cells and 2 reduction cells) 
	and the initial number of channels is 36.
	We train the network from scratch for 600 epochs with a batch size of 96. 
	The network is optimized via the SGD optimizer 
	with an initial learning rate of 0.025 (cosine annealing to 0), momentum of 0.9,
	weight decay of 3$e$-4, and gradient clipping at 5. 
	Cutout and drop-path with a rate of 0.2 are used for preventing overfitting. 
	
	\textbf{Main Results.}
	The evaluation results on CIFAR10/100 are shown in \tabref{tab:CIFAR10}. 
	DOTS only costs 0.26 GPU-days to achieve 97.51\% and 83.72\% accuracy on CIFAR10 and CIFAR100, respectively.
	Thanks to the decoupling of the operation and topology search, the number of candidate operations on edges is largely reduced, and both stages are fast to converge. DOTS improves DARTS by 0.51\% on CIFAR10 with lower search costs. The previous effort in gradient-based methods can be incorporated into the DOTS framework and further improved by the topology search, which is shown in \tabref{tab:improve_existing}.

	\begin{table*}[!tb]
		\centering
		\small
		\setlength{\tabcolsep}{2.5mm}
		\resizebox{0.8\linewidth}{!}{%
			\begin{tabular}{lc ccccc}
				\toprule[1pt]
				\multirow{2}{*}{\textbf{Architecture}} & \multicolumn{2}{c}{\textbf{Acc. (\%)}} & \multirow{2}{*}{\tabincell{c}{\textbf{Params}\\\textbf{(M)}}} & \multirow{2}{*}{\tabincell{c}{\textbf{Multi-Add}\\\textbf{(M)}}} & \multirow{2}{*}{\tabincell{c}{\textbf{Search Cost}\\\textbf{(GPU-days)}}} & \multirow{2}{*}{\textbf{Search Method}} \\ \cline{2-3}
				& \textbf{top-1} & \textbf{top-5} &  &  &  &  \\ \midrule[0.5pt]
				Inception-v1 \cite{szegedy2015going} &  69.8 & 89.9  & 6.6  &  1448  & N/A & N/A  \\
				MobileNet \cite{howard2017mobilenets} &  70.6 & 89.5  & 4.2  &  569  & N/A & N/A  \\
				ShuffleNet 2$\times$ (v1) \cite{zhang2018shufflenet} &  73.6 & 89.8  & 5.4  &  524  & N/A & N/A  \\
				ShuffleNet 2$\times$ (v2) \cite{ma2018shufflenet} &  74.9 & 92.4  & 7.4  &  591  & N/A & N/A  \\ \midrule[0.5pt]
				NASNet-A \cite{zoph2018learning} &  74.0  &  91.6  & 5.3  & 564 & 1800  & RL  \\
				AmoebaNet-C \cite{real2019regularized} &  75.7  &  92.4  & 6.4  & 570 & 3150  & EA  \\
				PNAS \cite{liu2018progressive} &  74.2  &  91.9  & 5.1  & 588 & 225  & SMBO  \\
				MnasNet-92 \cite{tan2019mnasnet} &  74.8  &  92.0  & 4.4  & 388 & 1667  & RL  \\ \midrule[0.5pt]
				DARTS (2nd order) (CIFAR10) \cite{liu2018darts} &  73.3  &  91.3  & 4.7  & 574 & 4.0  & GD \\
				SNAS (CIFAR10) \cite{xie2018snas} &  72.7  &  90.8  & 4.3  & 522 & 1.5  & GD \\
				P-DARTS (CIFAR10) \cite{chen2019progressive} &  75.6  &  92.6  & 4.9  & 557 & 0.3  & GD  \\
				GDAS (CIFAR10) \cite{dong2019searching} &  74.0   & 91.5  & 5.3 & 581  & 0.2   & GD  \\
				FairDARTS (CIFAR10) \cite{chu2019fair} &  75.1  &  92.5  & 4.8  & 541 & 0.4  & GD \\
				PC-DARTS (CIFAR10) \cite{xu2020pcdarts} &  74.9  &  92.2  & 5.3  & 586 & 0.1  & GD  \\
				SDARTS-ADV (CIFAR10) \cite{chen2020stabilizing} &  74.8  &  92.2  & 5.4  & 594 & 1.3  & GD  \\
				DropNAS (CIFAR10) \cite{hong20drop} &  75.5  & 92.6  & 5.2  & 572 & 0.6  & GD \\
				ASAP (CIFAR10) \cite{noy2020asap} &  73.7  &  91.5  & 3.8  & 427 & 0.2  & GD \\
				DOTS (CIFAR10) &  75.7  &  92.6 & 5.2 & 581 & 0.2  & GD \\  \midrule[0.5pt] 					
				ProxylessNAS (ImageNet) \cite{cai2018proxylessnas} &  75.1  & 92.5  & 7.1  & 465 & 8.3  & GD \\ 
				FairDARTS (ImageNet) \cite{chu2019fair} &  75.6  & 92.6  & 4.3  & 440 & 3  & GD \\ 
				PC-DARTS (ImageNet) \cite{xu2020pcdarts} &  75.8  &  92.7  & 5.3  & 597 & 3.8  & GD \\ 
				DOTS (ImageNet) & 76.0  &  92.8  & 5.3  & 596 & 1.3 & GD \\ \bottomrule[1pt]
		\end{tabular}}
		\vspace{-.15in}
		\caption{Comparison with state-of-the-art models on ImageNet. CIFAR10 and ImageNet mean the cell architecture is searched on CIFAR10 or ImageNet. 
		}  
		\label{tab:ImageNet}
		\vspace{-.1in}
		
	\end{table*}

	\begin{table}[!tb]
		
		\subfloat[Ablation study of different topology parameterized strategy.]{
			\label{tab:tpstrategy}
			\resizebox{0.9\linewidth}{!}{%
				\begin{tabular}{c|c|cc}
					\toprule[1pt]
					\tabincell{c}{\textbf{Topology Parameterized}\\\textbf{Strategy}} & \tabincell{c}{\textbf{Edge Numbers}\\\textbf{Constraint}} & \tabincell{c}{\textbf{CIFAR10}\\\textbf{Test Acc. (\%)}} & \tabincell{c}{\textbf{CIFAR100}\\\textbf{Test Acc. (\%)}} \\ \midrule[0.5pt]
					PC-DARTS& 2 & 97.38  $\pm$0.09 & 82.98  \\
					DOTS & 2 & 97.51 $\pm$0.06 & 83.72  \\ 
					\midrule
					Edge-Level Sigmoid & arbitrary & 97.26 $\pm$0.14 & 81.02\\
					DOTS & arbitrary & 97.53 $\pm$0.08 & 83.92\\ \bottomrule[1pt]
			\end{tabular}}
		}
		\\
		\subfloat[Ablation study of different operation search strategies. \#op: The numbers of operation retained on each edge.]{
			\label{tab:opstrategy}
			\resizebox{0.9\linewidth}{!}{
				\begin{tabular}{c|c|ccc}
					\toprule[1pt]
					\tabincell{c}{\textbf{Operation Search}\\\textbf{Strategy}} & \textbf{\#op} & \tabincell{c}{\textbf{CIFAR10}\\\textbf{Test Acc.  (\%)}} & \tabincell{c}{\textbf{CIFAR100}\\\textbf{Test Acc.  (\%)}} & \tabincell{c}{\textbf{Search Cost}\\\textbf{(GPU-days)}}\\ \midrule[0.5pt]
					DARTS-Top1 & 1 & 97.40 $\pm$0.09 & 83.07  & 0.22\\ 
					DARTS-Top2 & 2 & 97.42 $\pm$0.11 & 82.96  & 0.26\\ \midrule[0.5pt]
					Group-V1 \cite{li2019stacnas} & 4 & 97.48 $\pm$0.11 & 83.55  & 0.35 \\ 
					Group-V2 \cite{hong20drop} & 2 & 97.51 $\pm$0.06 & 83.72  & 0.26 \\ 
					\bottomrule[1pt]
				\end{tabular}
			}
		}
		\vspace{-.1in}
		\caption{Ablation studies of different strategies.}
		\label{tab:abaltion}
		\vspace{-.2in}
	\end{table}
	
	\subsection{Evaluation on ImageNet}
	
	\textbf{Search Settings.} 
	The whole search process takes 70 epochs, 30 epochs for the operation search and 40 epochs for the topology search. 
	The network is composed of 14 cells for both search stages. 
	The initial temperature $T_0$ is set to 10 and decay to 0.02 in the topology search. 
	The whole search process costs 7.8 hours (1.3 GPU-Days) on four NVIDIA Quadro RTX 8000 GPUs. Other search settings we keep the same as PC-DARTS \cite{xu2020pcdarts}, which can be found in supplementary.
	
	\textbf{Evaluation Settings.}
	The evaluation follows the mobile setting, where the input image size is set to $224\times 224$, 
	and the number of multiply-add operations is restricted to be fewer than 600M. 
	The network consists of 14 cells (12 normal cells and 2 reduction cells) 
	with an initial number of channels of 46.
	We train the network from scratch for 250 epochs with a batch size of 1024. 
	The SGD optimizer with an initial learning rate of 0.5 (warm up in the first 5 epochs and cosine annealing to 0), momentum of 0.9, and weight decay of 3$e$-5 is used.
	Additional enhancements follow P-DARTS \cite{chen2019progressive} and PC-DARTS \cite{xu2020pcdarts}, including label smoothing and an auxiliary loss tower.
	
	\textbf{Main Results.}
	The evaluation results are summarized in \tabref{tab:ImageNet}. 
	Most gradient-based methods search on a proxy dataset, \textit{e.g.}, CIFAR10, 
	and transfer the searched cell to ImageNet because their search cost on ImageNet is prohibitive. 
	While DOTS can search on ImageNet proxylessly, and requires the least search costs (1.3 GPU-days). 
	DOTS improves DARTS by 2.7\% of top-1 accuracy on ImageNet.

	\begin{figure*}[!tb]
		\centering
		\begin{minipage}[h]{0.5\linewidth}
			\subfloat[Operation searched results]{
				\label{tab:dartspolicy}
				\resizebox{0.98\linewidth}{!}{
					\begin{tabular}{cccccccc}
						\toprule[1pt]
						\textbf{EdgeID}     & 1      & 2      & 3      & 4      & 5      & 6      & 7  \\ \midrule[0.5pt] 
						\textbf{OP} & SkipC & Sep3x3 & SkipC & SkipC & Sep3x3 & SkipC & SkipC \\ \midrule[0.5pt]
						\textbf{EdgeID}     & 8      & 9      & 10     & 11     & 12     & 13     & 14     \\ \midrule[0.5pt] 
						\textbf{OP} & Sep3x3 & Sep3x3 & SkipC & Sep3x3 & Sep3x3 & Sep3x3 & Sep3x3 \\ \bottomrule[1pt]
					\end{tabular}
			}}\\
			\subfloat[DOTS topology search]{
				\includegraphics[width=\linewidth]{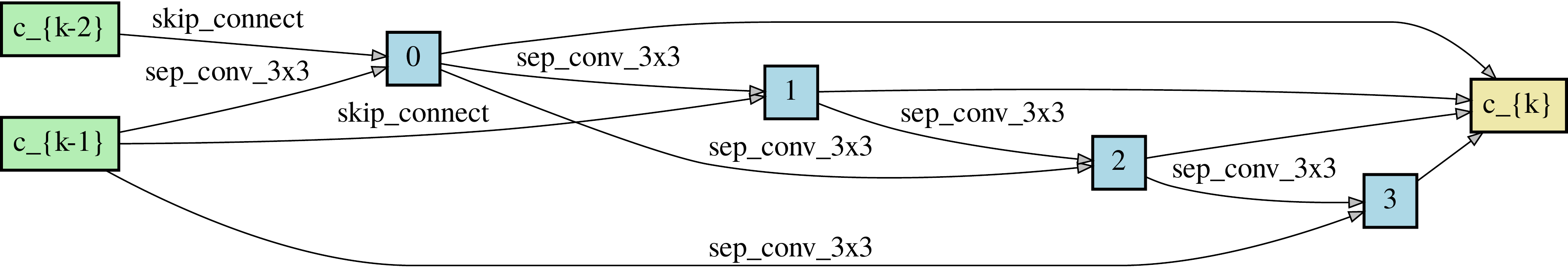}
				\label{fig:dotspolicy}	
			}
		\end{minipage}
		\hfill
		\begin{minipage}[h]{0.38\linewidth}
			\subfloat[DARTS' policy]{
				\includegraphics[width=0.86\linewidth,height=.60\linewidth]{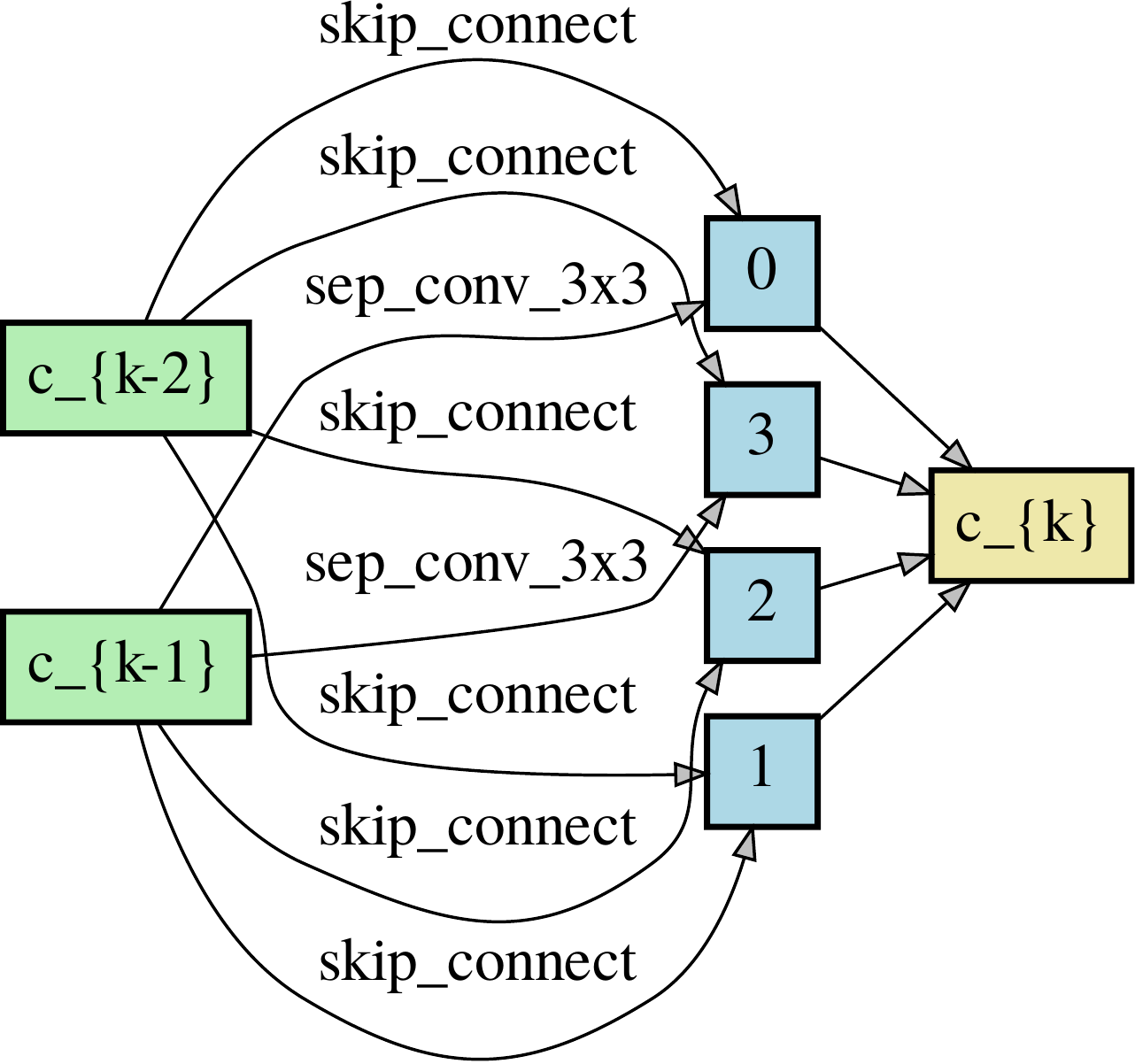}
				\label{fig:dartspolicy}	
			}
		\end{minipage}
		\caption{Results of different topology derivations. (a) The operations searched by DARTS on CIFAR10. (b) Deriving topology based on proposed topology search. (c) Deriving topology based on DARTS' policy.}  
		\label{fig:example}
	\end{figure*}
	
	\begin{figure}[!tb] \footnotesize
		\centering
		
		\subfloat[Operation searched results]{
			\label{tab:group_table}
			\resizebox{0.78\linewidth}{!}{
				\begin{tabular}{cccccccc}
					\toprule[1pt]
					\textbf{EdgeID}     & 1      & 2      & 3      & 4      & 5      & 6      & 7  \\ \midrule[0.5pt] 
					\textbf{OP Group 1} & SkipC & SkipC & SkipC & SkipC & SkipC & SkipC & SkipC \\
					\textbf{OP Group 2} & Dil3x3 & Sep3x3 & Sep3x3 & Sep3x3 & Sep3x3 & Sep3x3 & Sep3x3 \\ \midrule[0.5pt]
					\textbf{EdgeID}     & 8      & 9      & 10     & 11     & 12     & 13     & 14     \\\midrule[0.5pt]
					\textbf{OP Group 1} & SkipC & SkipC & SkipC & SkipC & SkipC & SkipC & Zero \\  
					\textbf{OP Group 2} & Dil3x3 & Dil3x3 & Dil5x5 & Sep3x3 & Sep3x3 & Dil3x3 & Sep3x3 \\ \bottomrule[1pt]
				\end{tabular}}
		}
		\\
		\subfloat[Topology searched results]{
			\includegraphics[width=0.78\linewidth]{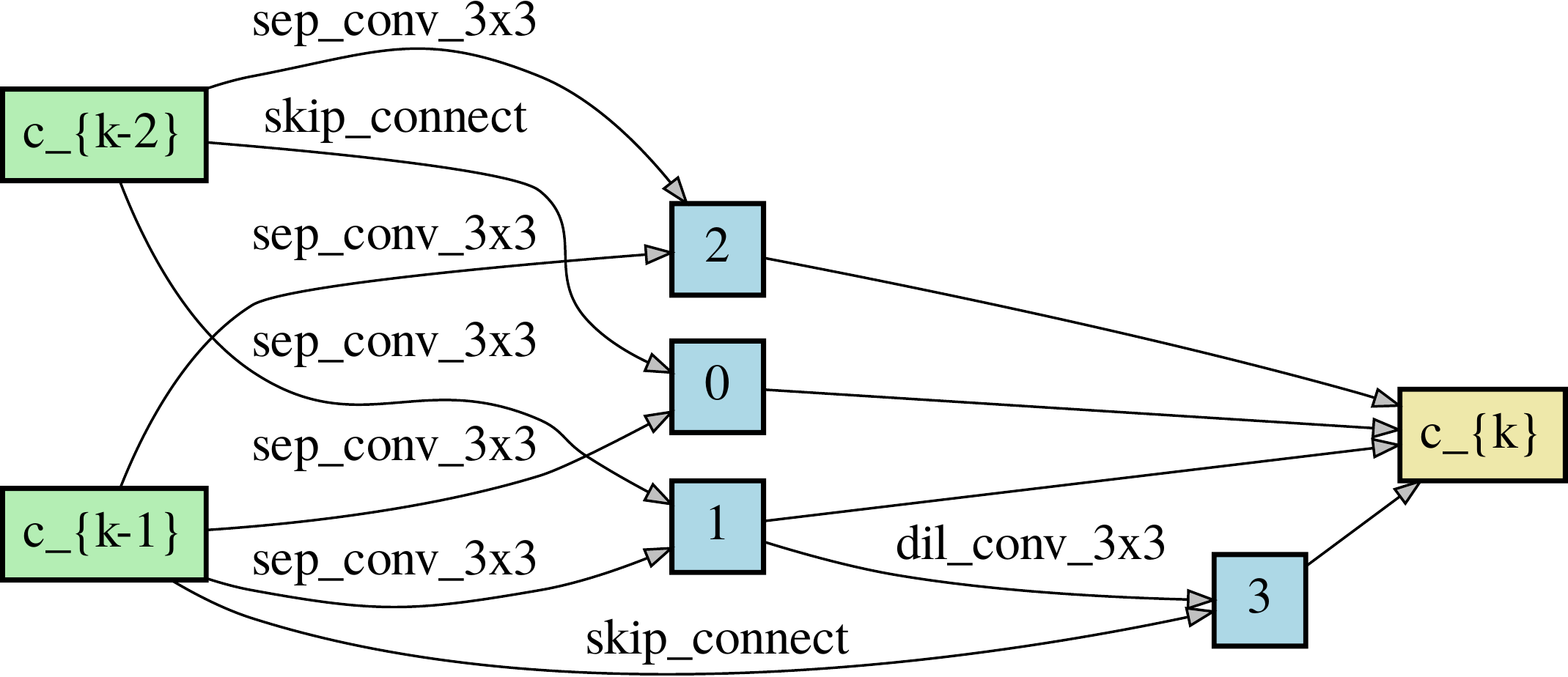}
			\label{fig:group_fig}	
		}		\vspace{-.1in}
		\caption{Results of the operation and topology search based on Group-V2 strategy.}  
		\label{fig:groupexample}
	\end{figure}

	\subsection{Ablation Study}
	\label{sec:ablation} 
	
	\textbf{Improving gradient-based NAS methods.}
	We incorporate five gradient-based methods into DOTS and validate the effectiveness of the topology search. The baseline is to derive the topology of the searched cell by DARTS policy. 
	Comparison results are summarized in \tabref{tab:improve_existing}. We can observe that adding topology search results in a consistent improvement over the baseline, verifying the effectiveness of the proposed topology search.
	An illustrative example is shown in \figref{fig:example}.
	We use DARTS for the operation search and the operation search results are shown in \figref{tab:dartspolicy}. Directly deriving its topology based on the operation weights, \textit{i.e.}, DARTS' policy, results in the searched cell in \figref{fig:dartspolicy}. We can find it is dominated by skip-connection and only achieves 80.74\% accuracy on CIFAR100. 
	\figref{fig:dotspolicy} shows the topology search results based on the same searched operations, which obtains a 2.33\% improvement (83.07\% \textit{vs.} 80.74\%). 
	We can find the topology search helps remedy the unstable results of the operation search by further choosing edge connections.
	
	\textbf{Influence of topology parameterized strategy.}
	We validate the effectiveness of the proposed edge combination weights and compare it with some other topology parameterized strategies.
	First, we compare with PC-DARTS that directly adds learnable weight on edges.
	Since it is not easy for the PC-DARTS strategy to support flexible edge numbers in the searched cell, we restrict both methods to retain two edges per node after the topology search.
	From the results in \tabref{tab:tpstrategy}, the proposed DOTS improves the edge weight of PC-DARTS by 0.13\% on CIFAR10. This mainly because the original purpose of the edge weight in PC-DARTS is to stabilize the training, not for the topology search. In comparison, the proposed edge combination weights can directly reflect the objective of edge selection.
	
	Then we validate the effectiveness of DOTS in arbitrary edge numbers setting. By removing the edge number constraint, the performance of DOTS is promoted from 83.72\% to 83.92\% on CIFAR100. 
	In the arbitrary edge numbers setting, a straightforward method is to replace the softmax activation with sigmoid activation on the edge weights and binarize the choice with a threshold. We name this strategy  edge-level sigmoid. From \tabref{tab:tpstrategy}, the DOTS has a clear advantage over the edge-level sigmoid strategy.
	
	\textbf{Influence of the operation search strategy.}
	We investigate the different operation search strategies and discuss their influence to the topology search.
	The baseline is DARTS-Top1, which retains one strongest operation on each edge for the topology search. 
	The operation search with Group-V2 strategy (details in supplementary) improves DARTS-Top1 by 0.11\% and 0.65\% on CIFAR10 and CIFAR100, respectively. The reason is DARTS-Top1 overlooks a case that some operations are topology-related. Pruning these operations in the operation search stage will eliminate the potential topology choices, resulting in suboptimal solutions.
	
	An illustrative example of the group strategy is shown in \figref{fig:groupexample}. From \figref{tab:group_table}, the best operation in the topology-related and topology-agnostic group is retained for the topology search. 
	\figref{fig:group_fig} shows the topology search results based on the operation search results in \figref{tab:group_table}.
	Direct retaining more operations, \textit{i.e.}, DARTS-Top2 has no obvious improvement, because it does not guarantee topology-related operations are preserved. Adding more groups in topology-agnostic operations \textit{i.e.}, Group-V1 strategy has no improvement but increases the search cost. 
	From this experiment, we can find preserve two operations (one topology-related and one topology-agnostic) is sufficient for the topology search.

	\section{Conclusion}
	In this paper, we study the topology derivation of existing gradient-based methods based on DARTS. The edge importance of DARTS is based on the operation weights, which can not correctly rank the stand-alone models with different topologies. Thus we propose to decouple the topology representation from the operation weights and make an explicit topology search. The proposed topology representation, \textit{i.e.}, edge combination weights, leads to the correct topology rating and supports flexible edge numbers.
	Apart from decoupling the operation and topology representation,
	we propose to decouple their searching processes to make an efficient and accurate topology search.
	Experiments on CIFAR and ImageNet demonstrate DOTS is an efficient and effective solution to differentiable NAS.
	
	\paragraph{Acknowledgement}
	
	This research was supported by the Major Project for New Generation of AI 
	under Grant No. 2018AAA0100400, 
	S\&T innovation project from Chinese Ministry of Education,
	and NSFC (61922046). 
	We thank MindSpore for the partial support of this work.
	
	{\small
		\bibliographystyle{ieee_fullname}
		\bibliography{DOTS}
	}
	
\end{document}